\documentclass{article}

\usepackage{PRIMEarxiv}

\usepackage[utf8]{inputenc} 
\usepackage[T1]{fontenc}    
\usepackage{hyperref}       
\usepackage{url}            
\usepackage{booktabs}       
\usepackage{amsfonts}       
\usepackage{nicefrac}       
\usepackage{microtype}      
\usepackage{lipsum}
\usepackage{fancyhdr}       
\usepackage{graphicx}       
\graphicspath{{media/}}     

\usepackage{float}
\usepackage{fixltx2e}
\usepackage{subfig}
\usepackage{array, booktabs, makecell}
\usepackage{multirow}
\usepackage{adjustbox}
\usepackage{chngcntr}
\usepackage{color,soul}

\newcommand{\sig}{\textsuperscript{$\dagger$}\hspace{0.0ex}}

\sethlcolor{yellow} 
\renewcommand\hl[1]{#1}

\title{Detect-and-Segment: a Deep Learning Approach
to Automate Wound Image Segmentation
}

\author{\large Gaetano Scebba\thanks{Mobile Health Systems Lab, Institute of Robotics and Intelligent Systems, Department of Health Sciences and Technology, ETH Zurich, Switzerland (e-mails: gae.scebba@gmail.com, walter.karlen@ieee.org)} , 
Jia Zhang\footnotemark[1] , 
Sabrina Catanzaro\thanks{Unit for Clinical and Applied Research, Balgrist University Hospital, Zurich, Switzerland} , 
Carina Mihai\thanks{Department of Rheumatology, University Hospital Zurich, University of Zurich, Switzerland} , 
\\ \large \textbf{Oliver Distler\footnotemark[3],
Martin Berli\thanks{Division of Technical Orthopedics, Department of Orthopedic Surgery, Balgrist University Hospital, Zurich, Switzerland} , 
Walter Karlen\footnotemark[1]}}

\begin{document}
\maketitle

\begin{abstract}
        Chronic wounds significantly impact quality of life. If not properly managed, they can severely deteriorate. Image-based wound analysis could aid in objectively assessing the wound status by quantifying important features that are related to healing. However, the high heterogeneity of the wound types, image background composition, and capturing conditions challenge the robust segmentation of wound images. We present Detect-and-Segment (DS), a deep learning approach to produce wound segmentation maps with high generalization capabilities. In our approach, dedicated deep neural networks detected the wound position, isolated the wound from the uninformative background, and computed the wound segmentation map. We evaluated this approach using one data set with images of diabetic foot ulcers. For further testing, 4 supplemental independent data sets with larger variety of wound types from different body locations were used.  The Matthews’ correlation coefficient (MCC) improved from  0.29 when computing the segmentation on the full image to 0.85 when combining detection and segmentation in the same approach. When tested on the wound images drawn from the supplemental data sets, the DS approach increased  the mean MCC from 0.17 to 0.85. Furthermore, the DS approach enabled the training of segmentation models with up to 90\% less training data while maintaining the segmentation performance. 
\end{abstract}


	\section{Introduction}
		\label{sec:introduction}
		Chronic wounds cause a great reduction of quality of life in patients \cite{lindholm2016}. If not 
		treated, they can lead to severe complications such as limb amputations and death \cite{escandon2011, chandan2019}. In fact, periodic examination and thorough treatment are essential to prevent deterioration \cite{han2017chronic}. Wound management requires to regularly assess, document, and treat the wound by medical professionals \cite{Othman2012}. Trained nurses annotate the relevant information in assessment reports that are then used to track the healing progression and plan the most appropriate treatment strategy. The assessment is typically performed by visual inspection of wound features, such as area and depth measurements, and the annotation of growth tissue \cite{khalil2014implementation}. Despite the importance of objective and accurate wound documentation, the assessment reports are often inconsistent and sparse~\cite{hess2005art}. 
		
		Wound imaging through cameras and smartphones has been adopted to support wound documentation with a more objective tool \cite{Wang2015f, Goyal2019, zhang2021wound}. Wound images are taken at each examination and medical professionals obtain insights on the wound healing progression by visually comparing with the images taken at past appointments. However, the use of wound images does not entirely eliminate the subjectivity of wound documentation and might even introduce an additional source of variability to the whole documentation procedure, due to the non-quantitative examination of past images.

		\begin{figure}
			\centering
			\includegraphics[width=0.6\linewidth]{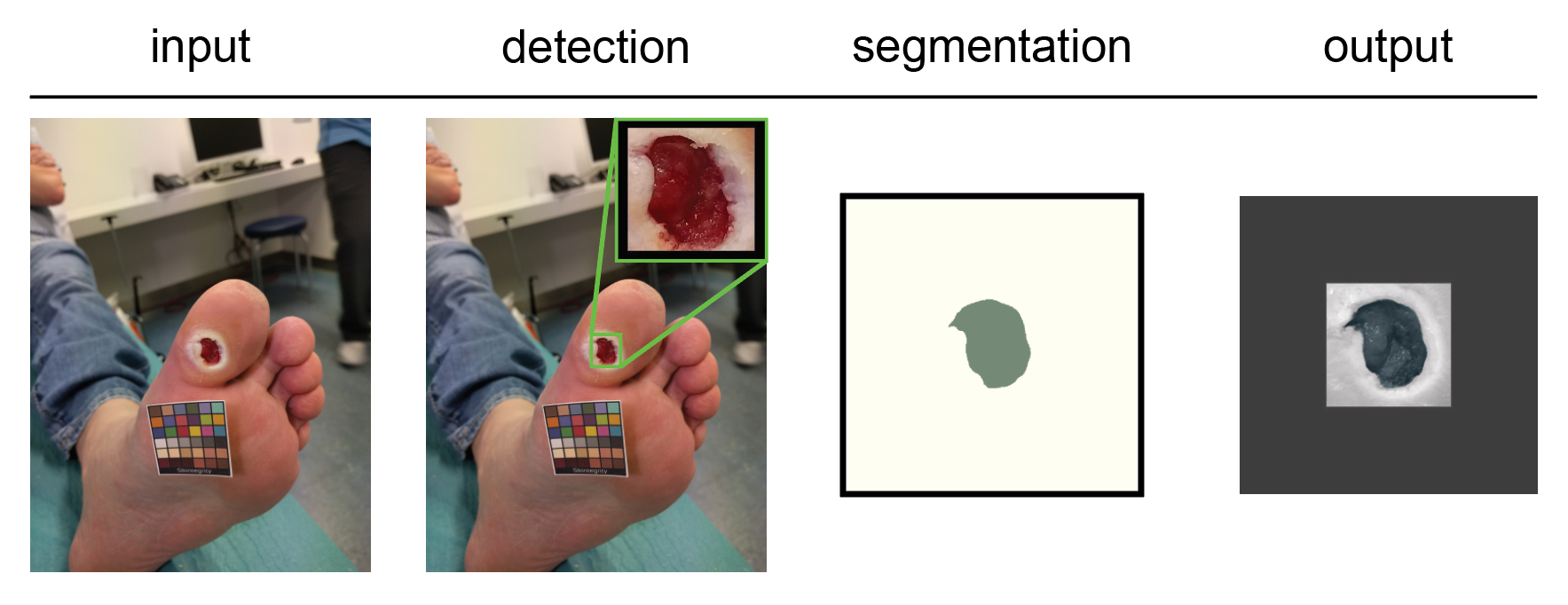}
			\caption{Proposed approach for robust segmentation of wound images. Dedicated deep neural networks detect the wound position, isolate the wound bed, and produce the wound segmentation map.}
			\label{fig:teaser}
		\end{figure}
			
	Image analysis techniques could potentially be used to obtain all the necessary information for the wound documentation in a more objective manner. In this process, a challenging task is the semantic segmentation. In this task,  each pixel of an image is classified to either belonging to the foreground or background class. In the case of wound images, the accurate distinction between wound bed and background information is important because the wound bed area is considered as one of the most relevant biomarkers for the prediction of wound healing \cite{sheehan2003} and excluding uninformative pixels can improve further analysis.
		
		However, generalizing segmentation performance across wound images is challenging. Heterogeneity of wound types, colors, shapes, body position, background composition, image capturing conditions, and capturing devices are all factors that lead to highly diverse wound images  \cite{lucas2020wound}. Clinical wound photography established a set of guidelines to standardize the working procedures and improve objectivity of the image capturing \cite{imi2019}. While these steps would provide wound nurses with additional workload, they do not consistently solve the generalization difficulties.
		
		The goal of this work was to simplify the wound imaging process with digital cameras by developing an image segmentation approach that is robust to the heterogeneity of wound images. We focused on the development of a deep learning approach to accomplish wound segmentation that was capable of high generalization towards out-of-distribution wound images. Our approach enabled a wound image segmentation without user interaction and consequently alleviated human efforts for image standardization during wound images collection (Fig.~\ref{fig:teaser}).

		\subsection{Deep learning in object detection and segmentation for medicine}
			Deep learning has been widely applied for biomedical image processing in pathology, radiology or dermatology. Numerous approaches proposed deep neural networks (DNN) to localize objects of interest such as lesions \cite{Goyal2019, dou2016} or anatomical structures \cite{liu2018, litjens2017}. Similarly, DNNs have also been adopted to segment single cells from electron microscopy sections \cite{ronneberger2015, lugagne2020} or from histopathological images \cite{xie2016b}, brain and cardiac anatomical substructures from magnetic resonance imaging (MRI) \cite{bernard2018} or computer tomography (CT) scans \cite{debrebisson2015}, tissue lesions from breast MRI \cite{moeskops2016}, lungs and liver CT scans \cite{harisson2017, hdensenet2018}, or skin lesions such as melanoma \cite{attia2017} and wounds \cite{Wang2015f} from smartphone images. However, the segmentation of medical images is often challenged by the highly variable background of many image domains. In some cases, pre-processing the images by removing uninformative background pixels, either manually or automatically, is a necessary step \cite{litjens2017}. In this work, we build on these advances to develop a novel approach to robust wound image segmentation.

		\subsection{Segmenting wound images}
			State of the art methods for wound image segmentation are based on several machine learning approaches, ranging from the use of classification from manually engineered image features, to the use of end-to-end DNNs. Initially, researchers have proposed to use ad-hoc image features, which can represent the complexity and variability of wound shapes and colors, in conjunction with supervised and unsupervised classifiers. These works have used color and texture features with region growing algorithm and optimal thresholding \cite{fauzi2018}, with support vector machines \cite{Wannous2007,wang2017twostage}, with neural networks \cite{veredas2010, song2012}, and with k-means clustering algorithm \cite{veredas2015, bhavani2018}. However, coping with the high variability of wounds is challenging and features that are designed and evaluated for a limited set of images might not be able to generalize and be effective to segment images with unseen type of wounds or different lighting conditions.
			
			To replace the need for manually designing features for wound image segmentation, research interest has also investigated the use of deep learning. Convolutional neural networks (CNN) have been used to perform image segmentation because they can autonomously learn to extract the most meaningful set of features for the segmentation task \cite{wang2020}. Wang \textit{et al.} proposed to use the ConvNet \cite{wang2015conv}, which is based on an encoder-decoder CNN architecture, and tested it on a data set containing diabetic foot ulcers that were manually pre-processed with a modified version of the GrabCut algorithm \cite{grabcut} to remove background. Similarly, other CNN architectures have been used, such as the U-Net \cite{ronneberger2015, Cui2019} or the fully convolutional network (FCN) \cite{Li2018, liu2017}, all in combination with pre-processing steps that remove uninformative background either by the manual interaction of the user \cite{Cui2019} or by manual features engineering to detect background pixels \cite{liu2018, liu2017}. Standardizing the background in advance, before taking the picture, has also been a technical choice proposed in several works to keep the background consistent over different wound images. These approaches used capturing box to capture wound feet pictures \cite{Wang2015f} or paper barriers to cover the background \cite{dorileo2010, Goyal2017}.
			
			In contrast to previous work that focused on the segmentation of diabetic foot images and developed semi-automatic methods which are dependent on user interaction, we present an automatic deep learning approach for the robust image segmentation of wounds. In this manuscript, we validate this approach and demonstrate robustness using 4 independent data sets with images from diverse wound types and different body locations. Our deep learning approach aims to address the challenge of high heterogeneity of wound images and enables the reduction of pixel-level training data with high preservation of segmentation performance.

	\section{Detect-and-segment approach}
\begin{figure}[th!]
			\centering
			\includegraphics[width=0.75\linewidth]{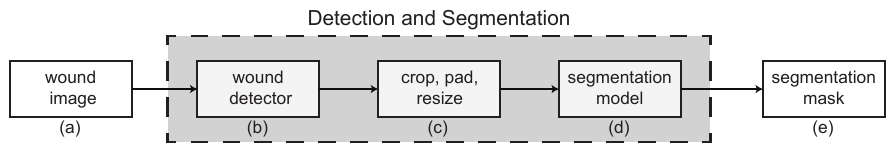}
			\caption{Detect-and-segment (DS) approach (grey box). The raw wound image (a) is processed with a wound detector neural network (b) to localize the wound within the image. The pre-processing module (c) performs cropping, zero-padding and resizing of the image to exclude uninformative background pixels. The segmentation model (d) produces the final segmentation mask (e). }
			\label{fig:algo}
		\end{figure}

			We have developed the detect-and-segment (DS), a deep learning approach for robust segmentation of wound images collected through either smartphones or digital cameras. Our approach consisted of three main steps (Fig~\ref{fig:algo}). First, the wound detector localized the wound bed region. Second, we processed the image using the localized wound bed to automatically remove uninformative background pixels. Third, a binary classification map was obtained by segmenting the resulting image into  wound bed and background pixels.
		
		\subsection{Wound bed detection}
			To localize the position of the wound bed region within the image, we trained an object \hl{detector}, with an architecture specifically designed for object detection. The model output was a set of bounding boxes that correspond to all the detected wounds in the image. Each box consisted of a set of 4 coordinates, which represented the top left and the bottom right corner of a box enclosing the wound bed. Additionally, the object \hl{detector} did output a confidence score that represented the probability for the detected object to be a wound. 
			{To select the best object detector for the DS approach, we conducted a systematic model selection. We optimized the object detection RetinaNet model \cite{retinanet2017} with three different feature extractors that were pre-trained on the Microsoft COCO data set \cite{coco2014}: resnet50 \cite{resnet2016}, Efficient Det \cite{efficientdet2019} with backbone D0 and D1, and the MobileNet \cite{mobilenet2017}. The training loss was set to the sum of the focal loss and the weighted L1 loss used for box regression \cite{retinanet2017}. We used the Adam optimizer \cite{adam2015} with 100 epochs training and an exponential learning rate decay strategy. Information regarding the selection of the wound detector is provided in appendix (Section A).}
	
		\subsection{Detection post-processing}
			In the post-processing stage, we selected only those bounding boxes with a confidence score greater than 0.3, which was obtained empirically. \hl{We noticed that the bounding box produced by the wound detector had the tendency to tightly enclose the wound, resulting in unbalanced pixel count between the wound and the background class. Therefore,} the area of the selected boxes was then increased by 50~\% to include part of the surrounding skin \hl{and reduce class imbalance between wound and background classes.} Thus, we cropped the wound image based on the coordinates of the post-processed box to exclude uninformative background pixels. To ensure standardized input size to the wound segmentation models, the obtained cropped image was zero padded 
			and resized to 512$\times$512 pixels, using the nearest neighbor interpolation algorithm provided by the Python Image Library \cite{pil2015}.
	
		\subsection{Wound bed segmentation}
			We considered the segmentation of the wound image as a binary segmentation task. Our goal was to produce a segmentation mask in which each value represented whether the correspondent pixel on the wound image belonged to the wound or background class.{We evaluated performance of the DS approach by training four different segmentation architectures: U-Net \cite{ronneberger2015}, ConvNet \cite{wang2015conv}, DeepLabV3 \cite{deeplab2017} with a ResNet-101 \cite{resnet2016} backbone, and a fully convolutional neural network (FCN) \cite{Goyal2017} with a VGG16 \cite{vgg2015} backbone. To account for the class imbalance between wound bed and background pixels of the wound images, we used the pixel-wise weighted binary cross entropy loss, where the weighting term was computed as the ratio between the total number of wound bed and background pixels of each training set fold. We used the Adam optimizer \cite{adam2015} for up to 100 epochs with an early stopping of 12 epochs on the validation set loss and an exponential learning rate decay strategy. To additionally combat overfitting risk, we augmented the training images by applying vertical and horizontal flip, brightness variation with maximum delta of 0.1, and scale and crop with a scale factor ranging from 0.1 to 2. }
	
	\section{Experiments}

		We evaluated the performance of the DS approach  using multiple independent data sets, including images from different body locations and with different wound types. To evaluate how the DS approach improves generalization, we extended the testing stage to out-of-distribution images from data sets not used for training. In addition, we conducted a series of experiments to test whether 1) the automatic detection subcomponent of the DS approach improves the wound segmentation performance compared to manual detection and absence of detection, 2) the combination of automatic detection and segmentation enhances the generalization capabilities of the segmentation models when tested on out-of-distribution wound images, and 3) the DS approach reduces the amount of segmentation masks needed for training stage.

		\subsection{Data sets}
			In this work, we relied on wound images obtained from six independent data sets. The SwissWOU data set was built by the authors for this manuscript and included two subsets of images with diabetic foot  and systemic sclerosis digital ulcers. 
			{The diabetic foot ulcers grand challenge 2020 (DFUC) \cite{cassidy2020, Goyal2019}}, Medtec \cite{medtec}, second healing intention (SIH) \cite{Yang2016}, and foot ulcer segmentation challenge (FUSC) \cite{wang2020} data sets  are publicly available data sets that included images with a variety of wound types. \hl{The data of the DFUC data set were used for the selection of the wound detector (Section A, appendix).} 
			The data of the SwissWOU subset with diabetic foot wounds were used during training and testing stages of the segmentation models. The data of the SwissWOU subset with systemic sclerosis digital ulcers, Medtec, SIH, and FUSC data sets were only used during testing.

			\paragraph{SwissWOU - diabetic foot ulcers (SW-DFU)} We collected wound images from March 2018 to January 2020 from 76 consenting patients (23 females and 53 males, mean age: 67~$\pm$~11 y) at the wound consultation unit of the Balgrist University Hospital, Zurich, Switzerland. The data set included  a total of 1096 high-resolution wound images from 756 wounds that were acquired with smartphone cameras. The images were collected at around 50 cm camera-wound distance, with the smartphone  oriented parallel to the wound plane. Wound nurses and research assistants were responsible for images collection. At each wound consultation appointment, we collected one image in ambient light and one image using an external LED ring flashlight attached to the smartphone.  

			\paragraph{SwissWOU – systemic sclerosis digital ulcers (SW-SSD)} We collected images from May 2019 to May 2020 from 15 consenting patients that conducted a regular, self-guided wound assessment at home as part o a study that investigated the wound image quality \cite{zhang2021wound}.  The patients were provided with a smartphone and a holder specifically designed to standardize the camera-wound distance. Wound images were automatically uploaded to a remote database using RedCap \cite{Harris2009}. Due to the small sizes of the digital  compared to diabetic foot ulcers, in the analysis of this work we included only those images with a wound coverage greater than 0.1 \%, which is defined as the ratio between the wound and image area. As a result, we included 7 patients (6 females and 1 male, mean age: 47~$\pm$~11 y) with a total of 63 digital ulcers images.  
	
			Image sizes of the SwissWOU data sets ranged from 1080$\times$1920 to 3000$\times$4000 pixels. The studies conducted to collect the images of the SW-DFU and SW-SSD data sets followed the ethical principles for research involving humans according the Declaration of Helsinki and were approved by the institutional ethics committee (ETH Zurich EK 2017-N-07 and 2019-N-22). The smartphone app used in both studies featured a camera module that disabled all the image processing techniques that might have altered image colors other than auto-exposure. The app was installed on a Nexus 5x (LG Electronics, Seoul, South Korea) and a Xiaomi Mi A1 (Xiaomi Corporation, Beijing, China).
	        
			\paragraph{Medtec} Medtec is a publicly available wound image database \cite{medtec} that includes a large variety of wound types. We used a subset of 53 randomly selected wound images containing wounds caused by pressure, surgery, diabetes, and meningitis, which were digitalized from 35~mm analog films. No further information regarding data collection was available.

			\paragraph{Secondary intention healing} The SIH data set was released by Yang \textit{et al.} \cite{Yang2016} and included wound images of patients undergoing secondary intention healing at the Severance Hospital, Seoul, South Korea. We randomly selected a subset of the data set consisting of 58 wound images from 12 patients, with diverse wound types from different body positions. The images of the data set were collected at 30-50~cm distance, with different digital single-lens reflex (DSLR) cameras and included wounds from different body locations such as face, hands, back, and feet. 
			
			\paragraph{Foot ulcer segmentation challenge} The FUSC data set included images of diabetic foot ulcers \cite{wang2020}. We randomly selected a subset 60 wound images. The data was collected by nurses and clinicians of the Advancing the Zenith of Healthcare Wound and Vascular Center, Milwaukee, Wisconsin, United States. The images were captured using a Canon SX 620 HS camera (Canon Inc., Tokyo, Japan) and an iPad Pro (Apple Inc., Cupertino, California, United States).
			
			\subsubsection{Annotation}
			The medical annotations of the SW-DFU, SW-SSD, Medtec, SIH, and FUSC data sets consisted of segmentation masks with pixel-level labels for each image, in which 1 corresponded to the wound bed and 0 to the background. The images of the FUSC data set included pixel-level labels of the wound bed produced by wound care specialists. Image labelling for the SW-DFU, SW-SSD, Medtec, and SIH data sets was performed by a research assistant and the first author, who were extensively trained by professional wound nurses to recognize the wound bed area. 

			\subsubsection{Data set split}

				The SW-DFU data set was used during training and testing stages \hl{of the segmentation models and was split considering the number of patients.} The SW-DFU data set included a deidentified patient id for each wound image that allowed a subject dependent split of the data set. 
				We split the \hl{SW-DFU} data set into training \hl{set} (90~\% of all patients), \hl{which was used for training and hyperparameter tuning}, and held-out test \hl{set} (10~\% of all patients), \hl{which was only used for final performance reporting}. The training set was additionally split according to the 5-fold cross validation scheme and each split was conducted at random stratification by number of images per patient and mean wound coverage (Table A.2, A.3, appendix). This way, we balanced all the folds and ensured that multiple wound images from the same patient were only included in one fold.

			\subsection{Hyperparameters tuning}
				To select the best set of hyperparameters, we followed a systematic approach with random search strategy. We performed a maximum of 50 hyperparameter optimization runs with hyperparameters randomly chosen from predefined ranges (Table~A.4, appendix). The performance of each set of hyperparameters was evaluated against the validation set following a 5-fold cross validation. The validation performances computed against each of the 5 validation sets were averaged to obtain one performance value per optimization run. We selected the best set of hyperparameters across the 50 optimization runs by choosing the \hl{segmentation} model with highest Matthews’ correlation coefficient. The best set of hyperparameters was then used to train \hl{each} model on the entire training set (merging training and validation set) and the final performance evaluated on the test set.
	
		\subsection{Evaluation}
			\subsubsection{Segmentation performance}
				To quantify the segmentation performance of the DS approach, we compared the performance of the UNet, ConvNet, DeepLab and FCN models, which were trained and tested only on the images of the SW-DFU that differed for the wound detection step:
				\begin{itemize}
					\item \textbf{Manual:} the images of train, validation, and test set were pre-cropped by manually selecting a rectangle that included the wound. 
					\item \textbf{None:} the images of train, validation, and test set were used with their full dimensions without any wound detection and cropping.
					\item \textbf{Automatic:} the images of train and validation set were manually pre-cropped and those of the test set were pre-cropped using the selected wound detection model.
				\end{itemize}

			\subsubsection{Generalizability}
\begin{figure}[t]
					\centering
					\includegraphics[width=0.6\linewidth]{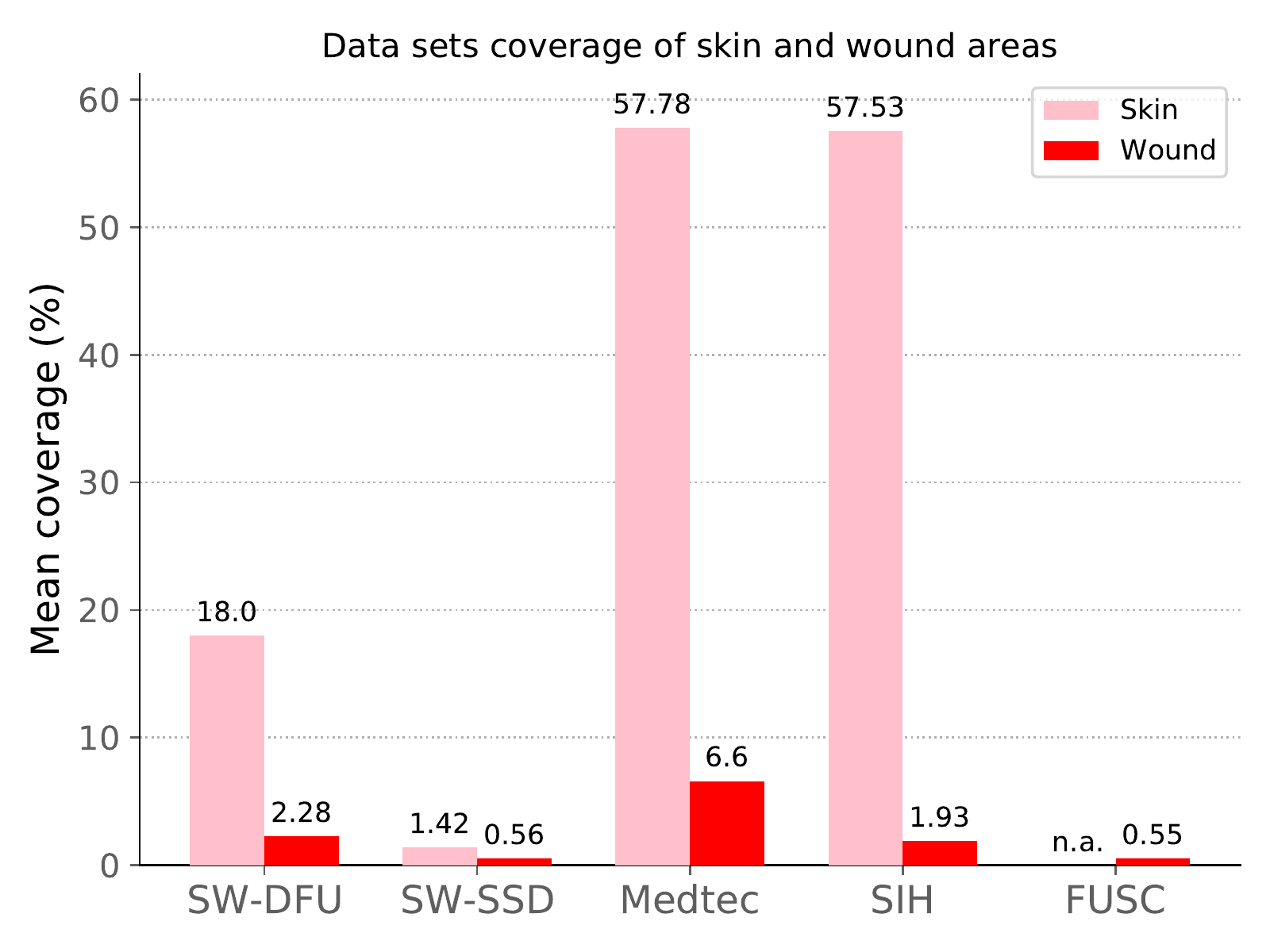}
					\caption{Data set mean coverage. Comparison of the skin and wound mean coverage percentage of the SwissWOU with diabetic foot ulcers (SW-DFU) and systemic sclerosis digital ulcers (SW-SSD), Medtec, SIH, and FUSC data sets. Mean skin coverage of the FUSC data set is not available as skin segmentation maps are not provided by the data set authors.}
					\label{fig:coverage}
				\end{figure}

				To evaluate the generalization capability of the DS approach, we computed its segmentation performance against out-of-distribution wound images. We compared the performance of all the segmentation models, which were trained only on the images of the SW-DFU data set and tested on the SW-SSD, Medtec, SIH, and FUSC data sets with and without Manual and Automatic wound detection step. These data sets not only included various wound types from different body locations, but also diverse background-foreground compositions with large differences in mean skin and wound coverage (Fig.~\ref{fig:coverage}). 
				
			\subsubsection{Impact of the data set size}
				To determine how the DS approach preserved the segmentation performance with reduced amount of training data, we re-trained all the segmentation models on the SW-DFU data set, by selecting their optimal hyperparameters and randomly removing from 10~\% up to 90~\% the training set images.

		\subsection{Performance metrics}	
			We evaluated the segmentation performance of the DS approach in terms of Matthews’ correlation coefficient (MCC), dice score and intersection over union (IoU, also known as Jaccard index). The MCC is widely used to handle data imbalance and can be seen as the discretization of the Pearson’s correlation coefficient for binary variables \cite{Boughorbel2017} and ranges from -1 to 1. Dice score and IoU are the most commonly used metrics in medical image segmentation because they adequately reflect the agreement for object segmentation \cite{Eelbode2020}. The IoU is defined as the ratio between the intersection portion of the detected and ground truth boxes divided by their union and ranges from 0 to 1.
			
			We additionally assessed whether the performance between segmentation models were statistically significant at a level of 95~\% significance. We first tested the normality of the analyzed MCC and IoU distributions using the Shapiro-~Wilk test and compared them with the Mann–Whitney U test. Bonferroni correction was applied in case of multiple comparisons. 

	\section{Results}
		\subsection{Segmentation performance}
			The DS approach achieved significantly ($p<0.05$) higher MCC and IoU than all the models trained on full dimension images without the detection step. Additionally, the DS approach significantly improved the segmentation performance of the U-Net and ConvNet compared to the manual detection step (Table \ref{tab:sw}).
			
			\begin{table}
			\caption{\textup{Segmentation Performance}. Comparison of UNet, ConvNet, DeepLab and FCN models trained on wound images pre-processed with automatic (DS approach), manual and no wound detector in terms of Matthews’ correlation coefficient (MCC) and intersection over union (IoU) for segmenting wound images from the SwissWOU subsets with diabetic feet (SW-DFU). In parentheses are the standard deviations. \sig = significant at $p < 0.05/3$ to performance of models trained on images pre-processed with the Automatic wound detector (DS approach).} 	
			\centering	
			\begin{small}
				\begin{adjustbox}{max width=\linewidth,center}

    				\begin{tabular}{llcc|cc|cc|cc}
    					\toprule
    					\multicolumn{2}{c}{} & \multicolumn{2}{c}{\textbf{UNet}} & \multicolumn{2}{c}{\textbf{ConvNet}} & \multicolumn{2}{c}{\textbf{DeepLab}} & \multicolumn{2}{c}{\textbf{FCN}}\\
    					\midrule
    					& Detection & MCC & IoU & MCC & IoU & MCC & IoU & MCC & IoU\\			
    					\midrule
    					\parbox[t]{2mm}{\multirow{3}{*}[0.9ex]{\rotatebox[origin=c]{90}{\textbf{SW-DFU}}}} 
    					&Manual 		      & \sig0.80 (0.04) & \sig0.67 (0.06) & \sig0.77 (0.05) & \sig0.63 (0.07) & \hspace{1ex}0.76 (0.06) & \hspace{1ex}0.61 (0.08) & \sig\textbf{0.81} (0.03) & \sig\textbf{0.68} (0.04)\\
    					&None & \sig0.29 (0.12) & \sig0.10 (0.07) & \sig0.35 (0.13) & \sig0.15 (0.10) & \sig0.60 (0.10) & \sig0.43 (0.08) & \sig0.59 (0.11) & \sig0.42 (0.08)\\
    					& Automatic 		  & 	\hspace{1ex}\textbf{0.85} (0.04) & \hspace{1ex}\textbf{0.75} (0.07) & \hspace{1ex}\textbf{0.82} (0.05) & \hspace{1ex}\textbf{0.70} (0.08) & \hspace{1ex}\textbf{0.78} (0.05) & \hspace{1ex}\textbf{0.63} (0.07) & \hspace{1ex}0.77 (0.07) & \hspace{1ex}0.63 (0.09)\\
    					
    					\toprule
    				\end{tabular}
				\end{adjustbox}

				\label{tab:sw}
			\end{small}
		\end{table}

		\subsection{Generalizability}
			The DS approach did not show significantly different performance ($p > 0.05$) compared to the manual detection step when tested on the images of the Medtec and SIH data sets for all the analyzed segmentation models (Table~\ref{tab:all}). The manual detection step showed significantly higher MCC and IoU than the DS approach ($p < 0.05$) when applied to the images of the SW-SSD subset for all the analyzed segmentation models (Table~\ref{tab:all}). Additionally, we found that the DS approach significantly outperformed the no-detection step ($p < 0.05$) for the U-Net and Convnet models when tested on the images of the SW-SSD, Medtec, SIH, and FUSC data sets (Table~\ref{tab:all}).
			\begin{table*}
			\caption{\textup{Quantitative generalization performance}. Comparison of UNet, ConvNet, DeepLab and FCN models trained on wound images pre-processed with automatic (DS approach) and manual wound detector in terms of Matthews’ correlation coefficient (MCC) and intersection over union (IoU) for segmenting wound images from the SwissWOU subsets with systemic sclerosis digital ulcers images (SW-SSD), Medtec, secondary intention healing (SIH), and foot ulcer segmentation challenge (FUSC) data sets. In parentheses are the standard deviations. \sig = significant at $p < 0.05$ to performance of models trained on images pre-processed with the Automatic wound detector (DS approach).} 	
			\centering	
			\begin{small}
    			\begin{adjustbox}{max width=\linewidth,center}

    				\begin{tabular}{llcc|cc|cc|cc}
    					\toprule
    					\multicolumn{2}{c}{} & \multicolumn{2}{c}{\textbf{UNet}} & \multicolumn{2}{c}{\textbf{ConvNet}} & \multicolumn{2}{c}{\textbf{DeepLab}} & \multicolumn{2}{c}{\textbf{FCN}}\\
    					\midrule
    					& Detection & MCC & IoU & MCC & IoU & MCC & IoU & MCC & IoU\\			
    					
    					\midrule
    					\parbox[t]{2mm}{\multirow{3}{*}[0.9ex]{\rotatebox[origin=c]{90}{\textbf{SW-SSD}}}} 
    					&	Manual & \sig\textbf{0.83} (0.02) & \sig\textbf{0.71} (0.03) & \sig\textbf{0.80} (0.03) & \sig\textbf{0.68} (0.05) & \sig\textbf{0.78} (0.08) & \sig\textbf{0.65} (0.11) & \sig\textbf{0.80} (0.09) & \sig\textbf{0.68} (0.10)\\			
    					&	None & \sig0.01 (0.01) & \sig0.00 (0.00) & \sig0.05 (0.08) & \sig0.01 (0.02) & \sig0.68 (0.09) & \sig0.51 (0.07) & \sig0.59 (0.05) & \sig0.41 (0.05)\\		
    					&Automatic & \hspace{1ex}0.74 (0.07) & \hspace{1ex}0.60 (0.09) & \hspace{1ex}0.74 (0.05) & \hspace{1ex}0.59 (0.06) & \hspace{1ex}0.75 (0.05) & \hspace{1ex}0.59 (0.07) & \hspace{1ex}0.67 (0.15) & \hspace{1ex}0.53 (0.13)\\
    					\midrule
    					\parbox[t]{2mm}{\multirow{3}{*}{\rotatebox[origin=c]{90}{\textbf{Medtec}}}} 
    					&	Manual & \hspace{1ex}0.90 (0.01) & \hspace{1ex}0.83 (0.02) & \sig\textbf{0.89} (0.01) & \sig\textbf{0.80} (0.02) & \hspace{1ex}0.88 (0.02) & \hspace{1ex}0.78 (0.04) & \hspace{1ex}0.85 (0.02) & \hspace{1ex}0.74 (0.03)\\
    					&	None & \sig0.11 (0.04) & \sig0.02 (0.02) & \sig0.44 (0.07) & \sig0.23 (0.05) & \sig0.83 (0.04) & \sig0.73 (0.06) & \sig0.78 (0.02) & \sig0.66 (0.03)\\			
    					&Automatic & \hspace{1ex}\textbf{0.90} (0.01) & \hspace{1ex}\textbf{0.83} (0.02) & \hspace{1ex}0.88 (0.01) & \hspace{1ex}0.79 (0.02) & \hspace{1ex}\textbf{0.89} (0.02) & \hspace{1ex}\textbf{0.80} (0.03) & \hspace{1ex}\textbf{0.85} (0.01) & \hspace{1ex}\textbf{0.74} (0.02)\\					
    					\midrule
    					\parbox[t]{2mm}{\multirow{3}{*}{\rotatebox[origin=c]{90}{\textbf{SIH}}}} 
    					&	Manual & \hspace{1ex}0.85 (0.02) & \hspace{1ex}0.78 (0.02) & \sig\textbf{0.87} (0.02) & \sig\textbf{0.79} (0.03) & \sig0.87 (0.02) & \sig0.80 (0.03) & \sig0.83 (0.02) & \hspace{1ex}0.75 (0.04)\\					
    					&	None & \sig0.25 (0.06) & \sig0.07 (0.02) & \sig0.53 (0.06) & \sig0.29 (0.06) & \sig\textbf{0.89} (0.02) & \sig\textbf{0.81} (0.04) & \sig\textbf{0.85} (0.03) & \hspace{1ex}\textbf{0.75} (0.05)\\			
    					&Automatic & \hspace{1ex}\textbf{0.87} (0.04) & \hspace{1ex}\textbf{0.81} (0.06) & \hspace{1ex}0.86 (0.04) & \hspace{1ex}0.79 (0.07) & \hspace{1ex}0.86 (0.03) & \hspace{1ex}0.80 (0.05) & \hspace{1ex}0.83 (0.04) & \hspace{1ex}0.76 (0.06)\\						
    
    					\midrule
    					\parbox[t]{2mm}{\multirow{3}{*}{\rotatebox[origin=c]{90}{\textbf{FUSC}}}} 
    					&	Manual & \hspace{1ex}\textbf{0.89} (0.01) & \hspace{1ex}\textbf{0.80} (0.03) & \hspace{1ex}0.87 (0.02) & \hspace{1ex}0.77 (0.03) & \sig0.81 (0.02) & \sig0.67 (0.04) & \sig0.80 (0.02) & \sig0.66 (0.03)\\
    					&	None & \sig0.22 (0.04) & \sig0.05 (0.02) & \sig0.34 (0.04) & \sig0.12 (0.03) & \sig0.75 (0.05) & \sig0.61 (0.08) & \sig0.80 (0.02) & \sig0.67 (0.02)\\
    					&   Automatic & \hspace{1ex}0.89 (0.02) & \hspace{1ex}0.79 (0.03) & \hspace{1ex}\textbf{0.88} (0.01) & \hspace{1ex}\textbf{0.78} (0.02) & \hspace{1ex}\textbf{0.83} (0.03) & \hspace{1ex}\textbf{0.69} (0.04) & \hspace{1ex}\textbf{0.82} (0.02) & \hspace{1ex}\textbf{0.69} (0.03)\\
    					\toprule
    				\end{tabular}
				\end{adjustbox}
				\label{tab:all}
			\end{small}
		\end{table*}

		\subsection{Impact of the data set size}
\begin{figure*}[hbt!]
			\centering
			\includegraphics[width=\linewidth]{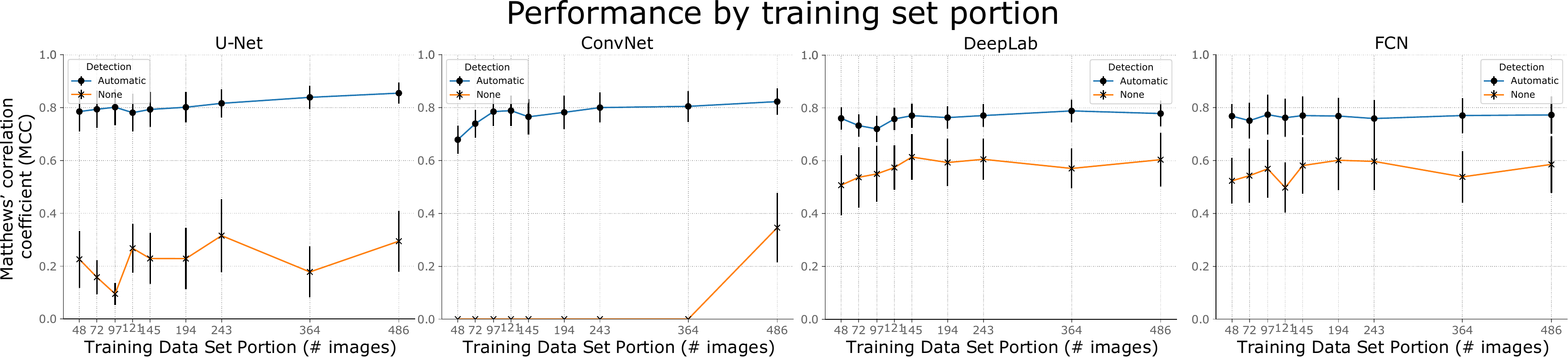}
			\caption{Performance comparison of the DS approach (dots, blue) and no-detection (x, orange) in terms of Matthews’ correlation coefficient (MCC, y-axis) when varying the amount of training set data (x-axis). The automatic detection of the DS approach enabled all the investigated segmentation models to preserve the performance when trained with reduced training set data.}
			\label{fig:ds}
		\end{figure*}

			In terms of segmentation performance as measured by the MCC, applying the DS approach allowed all the segmentation models to perform significantly better ($p~<~0.05$)  than the no-detection across a different amount of training data (Fig.~\ref{fig:ds}). In particular, when the DS approach was applied with U-Net and FCN models, they preserved the segmentation performance within the ranges of 6.4\% (MCC = [0.84 – 0.78] @ 90\% - 10\% of training set size) and 0.3\% (MCC = [0.770 - 0.768] @ 90\% - 10\% of training set size) across a reduction of 25 to 90\% the training set size.

	\section{Discussion}
		We presented and evaluated a novel deep learning approach for automatic segmentation of chronic wounds. Our approach featured two dedicated deep learning architectures that detect and segment the wound region without user interaction, in images collected from smartphones and DSLR cameras. To the best of our knowledge, this is the first study to extensively analyze the effects of the combination of the two processing steps, which have been only considered separately in past wound imaging literature. We evaluated the segmentation performance of the DS approach not only on the test set that was isolated from the same distribution of the training set, but also on wound images from independent data sets with different types of wounds. Our evaluation demonstrated that the DS approach enables the segmentation of wound images to be able to generalize well to images from out-of-distribution data, preserve performance with reduced training set size, and to be applied fully automatically.

		\subsection{Segmentation performance}
			The DS approach combined the tasks of detection and segmentation. The key element of the proposed approach was the automatic exclusion of uninformative background pixels given by the wound detection model. Pre-cropping the wound image was important for the training of segmentation models to focus only on learning to distinguish between the wound region and the surrounding skin. 
			
			We demonstrated that the DS approach improved the performance of all the analyzed segmentation models compared to the use of the full size images without pre-cropping. Furthermore, our results indicated that the automatic wound detection model of the DS approach enabled the segmentation models to perform similarly or even better than the use of manual wound detection. This confirms the importance of the wound detection task previously investigated in literature \cite{wang2015conv, Goyal2019} and expands the work of Goyal \textit{et al.} \cite{Goyal2017}, who successfully apply convolutional deep neural networks to wound images with normalized background. Together, these findings enable new applications where standardizing the background to make algorithms more effective and accurate is either not trivial (e.g. in a telemedicine application) or time consuming (e.g. in a clinical environment). 
		
		\subsection{Generalizability}
			We also evaluated the performance of the DS approach on wound images from multiple independent data sets. We found that training the segmentation models by only focusing on the wound-vs-skin problem provided all the models with improved generalization to images with different types of wounds and collected under different collection guidelines and environmental conditions. This was supported by the higher performance of all the models trained with both manual and automatic detection when tested on images with diabetic foot ulcers (SW-DFU, Table~\ref{tab:sw} and FUSC, Table~\ref{tab:all}), diabetic foot ulcers (SW-DFU, Table~\ref{tab:sw}), digital ulcers (SW-SSD, Table~\ref{tab:all}), and other wound types (Medtec, SIH, Table~\ref{tab:all}). However, when the number of images from a new distribution is not sufficient to train the wound detection models, as it occurred in the case of the digital ulcers (SW-SSD, Table~\ref{tab:all}), the performance of the segmentation models may be negatively influenced. In fact, we found that the manual detection on images with systemic sclerosis digital ulcers provided all segmentation models with higher segmentation performance than using the automatic detection (DS approach). To cope with lack of generalization due to low training data amount, more complex data augmentation techniques could be further explored to synthesize a variety of wounds (e.g. by using generative adversarial networks \cite{frid-adar2018}). In addition, research efforts in wound imaging should follow the progress made in dermoscopy imaging, in which large collections of several skin lesion images are publicly available \cite{isic2016, ham10000}. Wound imaging data sets should not only include images from diabetic foot wounds but also from other wound types, such as digital ulcers, post-surgical wounds, vascular ulcers, and burns.
			
\begin{figure*}
				\centering
				\includegraphics[width=\linewidth]{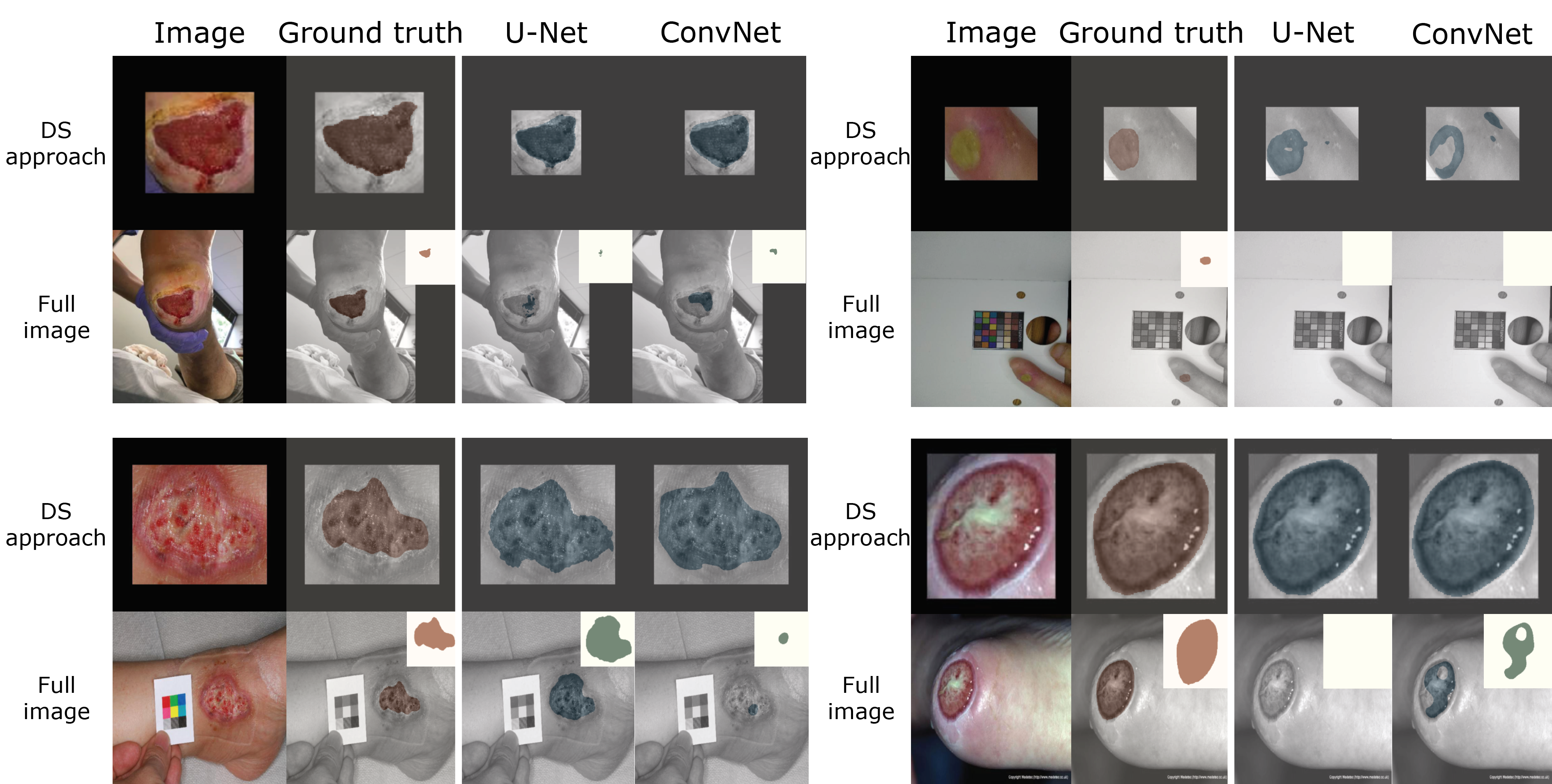}
				\caption{Qualitative generalization performance. Comparison of segmentation masks obtained with the DS approach and with the segmentation of the full image produced by U-Net and Convnet, from (top, left) the foot ulcer segmentation challenge (FUSC), (top, right) SwissWOU - systemic sclerosis digital ulcers (SW-SSD), (bottom, left) secondary healing intention (SIH), and (bottom, right) Medtec data sets. Top right white square within each of the images corresponding to full image row depicts the ground truth (brown) and computed (green) segmentation maps. The isolation of the wound bed from the uninformative background pixels performed by the automatic detection of the DS approach enables more robust and accurate segmentation compared to the segmentation of the full image. }
				\label{fig:qual}
			\end{figure*}

			Surprisingly, DeepLab and FCN models that were tested on the wound images of the SIH data set without detection showed higher segmentation performance than the DS approach (SIH, Table~\ref{tab:all}). The DeepLab and FCN models were built using backbone architectures that were pre-trained on larger non-medical data sets, while the U-Net and the ConvNet were trained from scratch using only wound images. This, in combination with the high amount of skin pixels of the SIH data set (Fig.~\ref{fig:coverage}) which results in a less heterogeneous background, could explain why the DeepLab and FCN models benefited from larger quantity of skin pixels around the wound. Nevertheless, considering the overall performance of the investigated segmentation models over all the analyzed data sets, applying the DS approach in combination with U-Net offered superior and more robust segmentation performance (Fig.~\ref{fig:qual}).
			
			Our findings on segmentation accuracy, reduced training data, and out-of-distribution testing complement the work of Wang \textit{et al.} \cite{wang2020}, who developed a fully automatic method for wound image segmentation. In combination, these results give the evidence that detecting and isolating the wound region from the background pixels is important for improving the accuracy of the segmentation and to generalize better to out-of-distribution wound images.
			
		\subsection{Impact of the data set size}
			The DS approach reduced the amount of training data for the segmentation models while preserving performance.  Using less segmentation maps for training implied less time needed by trained personnel to manually select all the pixels belonging to the wound region. 
			Despite reducing the number of segmentation maps, the labelling process for object detection was still needed. However, this process is in general faster because it only requires drawing a rectangle over the wound region. Therefore, it is a trade off between producing more labels for wound detection and less for segmentation in order to saving resources for manual data labelling. Moreover, future approaches could use unlabelled data with a semi-supervised learning approach such as the self-training \cite{xieselftraining2020}, which has been successfully applied for segmentation of medical images previously~\cite{zhangPancreaticDuctalSegm2020}.

		\subsection{Limitations}
			Despite the benefits of adopting an automatic wound detection step before the segmentation, cases of inaccurate wound detection must also be considered. The wound detector could produce bounding boxes with very large portions of the surrounding skin and even partial background or could lead to false positive and negative detections. The case of inaccurate but true positive detections could be mitigated by using bigger and richer data sets for the training of segmentation models. The risk of false positive detection could be solved by training the segmentation models with negative examples, such as images with region of healthy skin only or random background \cite{Goyal2017}. Furthermore, the case of false negative detection could be mitigated by introducing feedback mechanisms at the image collection time, which would notify the user to re-take the picture or change framing. 
			
			While we tested the performance of the DS approach with data from independent data sets, the study cohort of the SwissWOU data set used for training the segmentation models is likely not representative for all wound patients. It is evident that a validation study with larger and clinically representative cohort with multiple ethnicities and skin colors will be necessary to establish the performance of the DS approach for wound image segmentation. Additional clinical validation should also be conducted to evaluate whether the DS approach could be used to measure the wound bed area, which constitutes one of the most relevant parameters to assess the wound healing status and whether this process simplifies the clinical routines.
		
	\newpage
	\section{Conclusion}
		We presented a deep learning approach to improve generalization of wound image segmentation. We found that using the fusion of automated wound detection and segmentation improves the segmentation performance and enables the segmentation model to generalize well on out-of-distribution wound images. The DS approach allowed to preserve  segmentation performance when data with limited size is used for training to segment wound images. We demonstrated experimentally that our approach can be used to segment wound images from smartphones or professional digital cameras without the need of user interaction or background standardization during image collection. Our results confirm that machine learning based image analysis could  be implemented on mobile phones to support remote wound care in the future.
    
	\section*{Acknowledgements}
		We thank J. Toledo, S. Weidmann, L. Halter, and M. Emmenegger for assisting with patients recruitment and data collection, and M. Patwari, D. Chishima, and H. Orefice for technical developments of the wound image acquisition app.			

    \bibliographystyle{IEEEtran}
	\bibliography{wou_segm_thesis}
    
    \newpage
    \appendix
    \counterwithin{figure}{section}
    \counterwithin{table}{section}

    \section{Wound detector selection}
\subsection{Data set}
To identify the optimal wound detection model, we trained the model on the publicly available data set from the diabetic foot ulcers grand challenge (DFUC) 2020 \cite{cassidy2020, Goyal2019}. The DFUC data set consisted of 2000 wound images collected at the Lancshire Teaching Hospitals, United Kingdom. All images were acquired at a camera-wound distance of around 30-40, with parallel orientation of the camera to the plane of the wound and using different digital single-lens reflex (DSLR) cameras. Illumination with a flashlight was avoided and instead, adequate room lighting was adopted. Medical photographers with diabetic foot specialization were responsible for image collection. The images were already resized to 640$\times$480 pixels by applying the anti-alias down sampling filter implemented by the Python Imaging Library \cite{pil2015}. The medical annotations of the DFUC data set were produced by healthcare professionals specialized in diabetic foot ulcers and included of a set of four coordinates for each box enclosing the wound.
The DFUC data set additionally provided 2200 images for validation and testing purposes, which were not used for this work because they did not contain annotations. 

\subsubsection{Data set split}
The DFUC data set was split into training (90 \% of all wounds) and held-out test (10 \% of all wounds). To select the optimal hyperparameter configuration, the training set was additionally split using 5-fold cross validation scheme. To produce balanced sets across the folds, the splits were applied at random stratification by mean box coverage, which is defined as the ratio between the box and image areas (Table~\ref{tab:DFUC_split}). We used the DFUC data set to train the object detection models for the wound detection.

\subsection{Hyperparameters tuning}
 To choose the wound detector to be used in the DS approach for further analysis, we selected the object detector with the best performance on the SW-DFU test set. To select the best set of hyperparameters for the trained wound detectors, we performed a maximum of 50 hyperparameter optimization runs with hyperparameters randomly chosen from predefined ranges (Table~\ref{tab:hyperparameters}).

\subsection{Performance metrics}
The performance of each trained wound detector was evaluated in terms of mean average precision (mAP) at an intersection over union (IoU) of 50 \% (mAP @ 50\% IoU), a commonly used metric for object detection. The IoU is defined as the between the intersection portion of the detected and ground truth box divided by its union and ranges from 0 to 1. 

\subsection{Results}
The wound detector that best performed on the wound images of the SW-DFU data set was the MobileNet (Table~\ref{tab:det_performance}). We selected the MobileNet as wound detection model of the DS approach.

\begin{table*}[hbt!]
	\centering
	\caption{Split statistics for the DFUC Data set. Training, validation, and test fold statistics.}
	\begin{small}
		\begin{tabular}{ccccc}
			\toprule
			\thead{fold \\ (\#)} & \thead{wounds/image \\ train set (\#)} & \thead{mean coverage \\ train set (\%)} & \thead{wounds/image \\ val set (\#)}  & \thead{mean coverage \\ val set (\%)}  \\
			\midrule
			1 & 1796 & 4.38 & 450 & 5.01\\
			2 & 1797 & 4.55 & 449 & 4.30\\			
			3 & 1797 & 4.51 & 449 & 4.48\\
			4 & 1797 & 4.62 & 449 & 4.02\\						
			5 & 1797 & 4.46 & 449 & 4.70\\	
			\midrule		
			Test set & - & - & 250 & 4.37 \\ 
			\bottomrule
		\end{tabular}        
	\end{small}
	\label{tab:DFUC_split}
\end{table*}

\begin{table*}[hbt!]
	\centering
	\caption{Split statistics for the SwissWOU data set with diabetic foot ulcers (SW-DFU) data set - Segmentation performance experiment without detection step "\textit{None}". Training, validation, and test fold statistics.}	
	\begin{small}
		\begin{tabular}{ccccccc}
			\toprule
			\thead{fold \\ (\#)} & \thead{patients \\ train set (\#)} & \thead{images/patient \\ train set (\#)} & \thead{mean coverage \\ train set (\%)} & \thead{patients \\ val set (\#)} & \thead{images/patient \\ val set (\#)}  & \thead{mean coverage \\ val set (\%)}  \\
			\midrule
			1 & 54 & 8 & 2.28 & 13 & 10 & 1.47\\
			2 & 53 & 9 & 2.09 & 14 & 6 & 2.56\\			
			3 & 54 & 8 & 2.23 & 13 & 9 & 1.66\\
			4 & 53 & 8 & 2.20 & 14 & 10 & 1.84\\						
			5 & 54 & 9 & 1.89 & 13 & 7 & 3.07\\	
			\midrule		
			Test set & - & - & - & 7 & 8 & 3.10\\ 
			\bottomrule
		\end{tabular}        
	\end{small}

	\label{tab:sw-d full}
\end{table*}

\begin{table*}[hbt!]
	\centering
	\caption{Split statistics for the SwissWOU data set with diabetic foot ulcers (SW-DFU) data set - Segmentation performance experiment with manual detection step "\textit{Manual}". Training, validation, and test fold statistics.}	
	\begin{small}
		\begin{tabular}{ccccccc}
			\toprule
			\thead{fold \\ (\#)} & \thead{patients \\ train set (\#)} & \thead{images/patient \\ train set (\#)} & \thead{mean coverage \\ train set (\%)} & \thead{patients \\ val set (\#)} & \thead{images/patient \\ val set (\#)}  & \thead{mean coverage \\ val set (\%)}  \\
			\midrule
			1 & 54 & 9 & 13.15 & 13	& 9 & 10.29\\
			2 & 54 & 9 & 12.92 & 13	& 8	& 11.23\\			
			3 & 53 & 9 & 12.09 & 14	& 6	& 14.49\\
			4 & 53 & 8 & 12.45 & 14 & 9	& 13.17\\						
			5 & 54 & 8 & 12.35 & 13 & 10 & 13.61\\	
			\midrule		
			Test set & - & - & - & 7 & 8 & 10.77\\ 
			\bottomrule
		\end{tabular}        
	\end{small}
	\label{tab:sw-d full}
\end{table*}

\begin{table}
  \begin{minipage}{.5\linewidth}
  	\caption{Hyperparameters. \textup{Ranges used for hyperparameter optimisation of detection (top) and segmentation models (bottom). Parentheses indicate continuous ranges within the indicated limits sampled at uniform probability. Comma-delimited lists indicate discrete choices with equal selection probability.}}
    \begin{small}
		\begin{tabular}{l@{\hskip 2.5ex}l@{\hskip 5.5ex}r}
			\toprule
			& Hyperparameter & Range / Choices\\
			\midrule
			\parbox[t]{2mm}{\multirow{5}{*}{\rotatebox[origin=c]{90}{Detection}}}
			& Batch size  & 8, 16, 32, 64\\
			& L$2$ regularizer & (0.000001, 0.001)\\
			& Optimizer momentum & 0.99, 0.97, 0.95, 0.90\\
			& Initial learning rate & (0.000003, 0.003)\\
			& Learning rate decay & 0.97, 0.95, 0.93, 0.85\\
			\midrule
			\parbox[t]{2mm}{\multirow{6}{*}{\rotatebox[origin=c]{90}{Segmentation}}}&&\\
						& Batch size  & 4, 8, 16, 32\\
			& Weight decay & 0.0001, 0.00001, 0.0\\
			& Initial learning rate & (0.00003, 0.03)\\
			& Learning rate decay & 0.99, 0.97, 0.95, 0.93, 0.89\\
			&&\\
			\bottomrule
		\end{tabular}
	\end{small}
	\label{tab:hyperparameters}
  \end{minipage}
    \begin{minipage}{.5\linewidth}
	\caption{Wound detection performance. Comparison of Resnet50, EfficientDet-D0, EfficientDet-D1, and MobileNet models trained on the DFUC data set in terms of mean average precision (mAP) at 50\% intersection over union (IoU) for detecting wounds from images of the SwissWOU subsets with diabetic feet (SW-DFU).}
	\centering
	\begin{small}
		\begin{tabular}{l@{\hskip 15ex}c@{\hskip 3.8ex}}
			\toprule
			\thead{Detection \\ model} & \thead{mAP @ \\ 50\% IoU} \\
			\midrule
			resnet50 & {0.751}\\
			EfficientDet - D0 & {0.734} \\
			EfficientDet - D1 & {0.679} \\
			MobileNet & {\textbf{0.762}} \\
			\bottomrule
		\end{tabular}
	\end{small}
	\label{tab:det_performance}
  \end{minipage}%
\end{table}

\begin{figure}
    \centering
    \includegraphics[width=\linewidth]{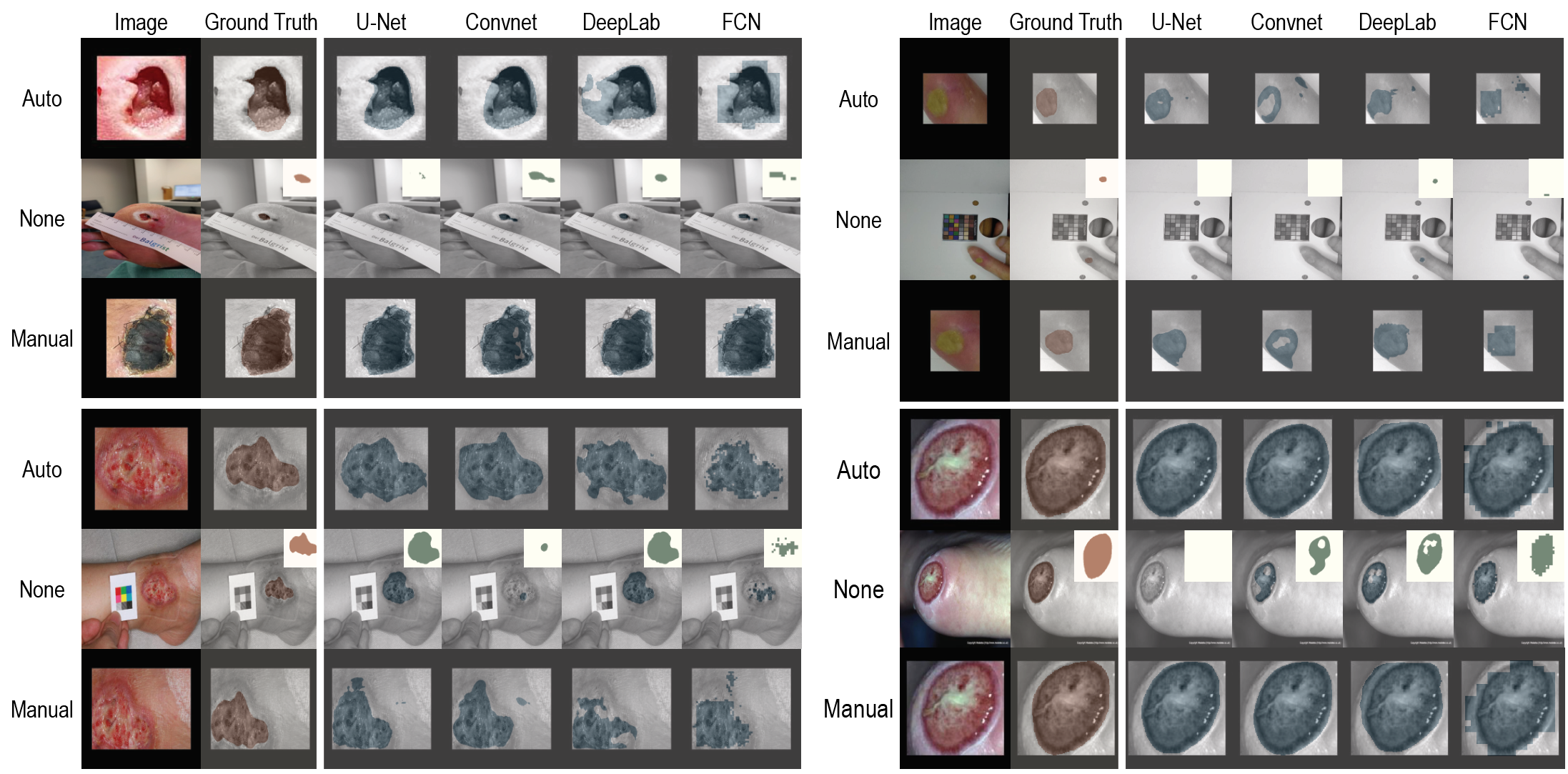}
    \caption{Qualitative generalization performance. Comparison of segmentation masks obtained with the DS approach and with the segmentation of the full image produced by U-Net, Convnet, DeepLab, and FCN from (top, left) the foot ulcer segmentation challenge (FUSC), (top, right) SwissWOU - systemic sclerosis digital ulcers (SW-SSD), (bottom, left) secondary healing intention (SIH), and (bottom, right) Medtec data sets. Top right white square within each of the images corresponding to full image row depicts the ground truth (brown) and computed (green) segmentation maps. The isolation of the wound bed from the uninformative background pixels performed by the automatic detection of the DS approach enables more robust and accurate segmentation compared to the segmentation of the full image.}
    \label{fig:my_label}
\end{figure}

\end{document}